\def\year{2018}\relax
\documentclass[letterpaper]{article} %DO NOT CHANGE THIS
\usepackage{aaai18}  %Required
\usepackage{times}  %Required
\usepackage{helvet}  %Required
\usepackage{courier}  %Required
\usepackage{url}  %Required
\usepackage{graphicx}  %Required
\usepackage{amsmath, amssymb}
\usepackage{txfonts}
\usepackage{booktabs}
\usepackage{comment}
\newtheorem{definition}{Definition}
\begin{document}
% The file aaai.sty is the style file for AAAI Press 
% proceedings, working notes, and technical reports.
%
\title{TorusE: Knowledge Graph Embedding on a Lie Group}

\author{Takuma Ebisu${}^\text{1,2}$\and Ryutaro Ichise${}^\text{2,1,3}$\\
	${}^\text{1}$SOKENDAI (The Graduate University for Advanced Studies)\\
	2-1-2 Hitotsubashi, Chiyoda-ku, Tokyo, Japan\\
	${}^\text{2}$National Institute of Informatics\\
	2-1-2 Hitotsubashi, Chiyoda-ku, Tokyo, Japan\\
	${}^\text{3}$National Institute of Advamced Industrial Science and Technology\\
	2-3-26 Aomi, Koto-ku, Tokyo, Japan\\
	\{takuma,ichise\}@nii.ac.jp\\
}

\maketitle
\begin{abstract}
Knowledge graphs are useful for many artificial intelligence (AI) tasks.
However, knowledge graphs often have missing facts.
To populate the graphs, knowledge graph embedding models have been developed.
Knowledge graph embedding models map entities and relations in a knowledge graph to a vector space and predict unknown triples by scoring candidate triples.
TransE is the first translation-based method and it is well known because of its simplicity and efficiency for knowledge graph completion.
It employs the principle that the differences between entity embeddings represent their relations.
The principle seems very simple, but it can effectively capture the rules of a knowledge graph.
However, TransE has a problem with its regularization.
TransE forces entity embeddings to be on a sphere in the embedding vector space. 
This regularization warps the embeddings and makes it difficult for them to fulfill the abovementioned principle.
The regularization also affects adversely the accuracies of the link predictions.
On the other hand, regularization is important because entity embeddings diverge by negative sampling without it.
This paper proposes a novel embedding model, TorusE, to solve the regularization problem.
The principle of TransE can be defined on any Lie group. A torus, which is one of the compact Lie groups, can be chosen for the embedding space to avoid regularization.
To the best of our knowledge, TorusE is the first model that embeds objects on other than a real or complex vector space, and this paper is the first to formally discuss the problem of regularization of TransE.
Our approach outperforms other state-of-the-art approaches such as TransE, DistMult and ComplEx on a standard link prediction task. We show that TorusE is scalable to large-size knowledge graphs and is faster than the original TransE.

\end{abstract}
\section{Introduction}
Knowledge graphs are one of the ways to describe facts of the real world in a form that a computer can easily process. 
Knowledge graphs such as YAGO \cite{DBLP:conf/www/SuchanekKW07}, DBpedia \cite{DBLP:conf/semweb/AuerBKLCI07} and Freebase \cite{Bollacker:2008:FCC:1376616.1376746} are used for many tasks, such as question answering, content tagging, fact checking, and knowledge inference. 
Although some knowledge graphs contain millions of entities and billions of facts, they still might be incomplete and have missing facts.
Hence, it is required to develop a system that can complete knowledge graphs automatically.

In a knowledge graph, facts are stored in the form of a directed graph.
Each node represents an entity in the real world and each edge represents the relation between entities. 
A fact is described by a triple $(h,r,t)$, where $h$ and $t$ are entities and $r$ is a relation directed from $h$ to $t$.
Some relations are strongly related.
For example, the relation \texttt{HasNationality} is related with the relation \texttt{CityOfBirth}.
Hence, if the triple $(\texttt{DonaldJohnTramp}, \texttt{HasNationality}, \texttt{U.S.})$ is not stored while $(\texttt{DonaldJohnTramp}, \texttt{CityOfBirth}, \texttt{NewYorkCity})$ is stored in a knowledge graph, the former can be easily predicted because most people born in New York City have the nationality of U.S.
Many kinds of models have been developed to predict unknown triples and to complete knowledge graphs through a link prediction task to predict the missing $h$ or $t$.

TransE, the original translation-based model for link prediction tasks, was proposed by Bordes et al. \shortcite{DBLP:conf/nips/BordesUGWY13} and it is well known because of its effectiveness and simplicity.
TransE embeds triples and relations on a real vector space with the principle $\boldsymbol{h}+\boldsymbol{r}=\boldsymbol{t}$, where $\boldsymbol{h}, \boldsymbol{r}$ and $\boldsymbol{t}$ are embeddings of $h, r$ and $t$, respectively, if the triple $(h,r,t)$ is stored in the knowledge graph used as training data.
Although it is very simple, the principle can capture the structure of a knowledge graph efficiently.
Many extended versions of TransE have been proposed. These include TransH \cite{DBLP:conf/aaai/WangZFC14}, TransG \cite{DBLP:conf/acl/0005HZ16} and pTransE \cite{DBLP:conf/emnlp/LinLLSRL15}.
On the other hand, various types of bilinear models, such as DistMult \cite{DBLP:journals/corr/YangYHGD14a}, HolE \cite{DBLP:conf/aaai/NickelRP16} and ComplEx \cite{DBLP:conf/icml/TrouillonWRGB16}, have been proposed recently and they achieve high accuracy on link prediction tasks with the metric HITS@\textit{1}.
The TransE model does not yield good results with the metric HITS@\textit{1},  but TransE is competitive with bilinear models with the metric HITS@\textit{10}. 
We find the reason for the TransE results is its regularization.
TransE forces entity embeddings to be on a sphere in the embedding vector space.
It conflicts with the principle of TransE and warps embeddings obtained by TransE.
In this way, it affects adversely the accuracies of the link predictions,
while it is required for TransE because embeddings diverge unlimitedly without it.

In this paper, we propose a model that does not require any regularization but has the same principle as TransE by embedding entities and relations on another embedding space, a torus.
Several characteristics are required for an embedding space to operate under the strategy of TransE.
A model under the strategy can actually be defined well on a Lie group of mathematical objects.
By choosing a compact Lie group as an embedding space, embeddings never diverge unlimitedly and regularization is no longer required.
Thus, we choose a torus, one of the compact Lie groups, for an embedding space and propose 
a novel model, TorusE. 
This approach allows the model to learn embeddings, which follow the TransE principle more precisely, and outperforms alternative approaches for link prediction tasks.
Moreover, TorusE is more scalable to large-size knowledge graphs because its complexity is the lowest compared with other methods, and we show that it is faster than TransE empirically because of the reduced calculation times without regularization.

The remainder of this paper is organized as follows.
In Section 2, we discuss related work for link prediction tasks.
In Section 3, we briefly introduce the original translation-based method, TransE, and mention its regularization flaw.
Then, the conditions required for an embedding space are analyzed to find another embedding space.
In Section 4, we propose a new approach to obtain embeddings by changing an embedding space to a torus.
This approach overcomes the regularization flaw of TransE.
In Section 5, we present an experimental study in which we compare our method with baseline results of benchmark datasets.
In Section 6, we conclude this paper.

\section{Related Work}
Various models have been proposed for knowledge graph completion through the link prediction task.
These models can be roughly classified into three types: translation-based models, bilinear models and neural network-based models. 
We describe notations here to discuss related work. 
$h$, $r$ and $t$ denote a head entity, relation, and a tail entity,
respectively. The bold letters $\boldsymbol{h}, \boldsymbol{r}$ and $\boldsymbol{t}$ denote embeddings of $h$, $r$ and $t$, respectively, on an embedding space $\mathbb{R}^n$.
$E$ and $R$ represent sets of entities and relations, respectively.
\subsection{Translation-based Models}
The first translation-based model is TransE \cite{DBLP:conf/nips/BordesUGWY13}.
It has gathered attention because of its effectiveness and simplicity.
TransE was inspired by the skip-gram model \cite{DBLP:journals/corr/abs-1301-3781,DBLP:conf/nips/MikolovSCCD13}, in which the differences between word embeddings often represent their relation.
Hence, TransE employs the principle $\boldsymbol{h}+\boldsymbol{r}= \boldsymbol{t}$.
This principle efficiently captures first-order rules such as  $``\forall e_1, e_2\!\!\in\!\! E,\: (e_1,r_1,e_2)\rightarrow (e_1,r_2,e_2)$", ``$\forall e_1, e_2\!\! \in\!\! E,\ (e_1,r_1,e_2)\rightarrow (e_2,r_2,e_1)$" and ``$\forall e_1, e_2\!\! \in\!\! E,\ \{\exists e_3 \!\!\in\!\! E, (e_1,r_1,e_3)\land(e_3,r_2,e_2)\}\rightarrow (e_2,r_3,e_1)$".
The first one is captured by optimizing embeddings so that $\boldsymbol{r}_1=\boldsymbol{r}_2$ holds, the second one is captured by optimizing embeddings so that $\boldsymbol{r}_1=-\boldsymbol{r}_2$ holds, and the third one is captured by optimizing embeddings so that $\boldsymbol{r}_1+\boldsymbol{r}_3=\boldsymbol{r}_2$ holds.
It was pointed by many researchers that the principle was not suitable to represent 1-N, N-1 and N-N relations.
Some models that extend TransE have been developed for solving those problems.

TransH \cite{DBLP:conf/aaai/WangZFC14} projects entities on the hyperplane corresponding to a relation between them.
Projection makes the model more flexible by choosing components of embeddings to represent each relation.
TransR \cite{DBLP:conf/aaai/LinLSLZ15} has a matrix for each relation and the entities are mapped by linear transformation that multiplies the matrix to calculate the score of a triple. 
TransR is considered as generalized TransH because projection is one of linear transformations.
These models have an advantage in power of expression comparing with TransE.
At the same time, however, they easily become overfitted.

TransE can be extended in other ways.
In TransG \cite{DBLP:conf/acl/0005HZ16}, a relation contained in a knowledge graph can have multiple meanings, and so a relation is represented as multiple vectors. 
pTransE \cite{DBLP:conf/emnlp/LinLLSRL15} takes relation paths between entities into account to calculate the score of a triple.
A relation path is represented by the summation of each relation in a path.

\subsection{Bilinear Models}
Recently, bilinear models have yielded great results of link prediction.
RESCAL \cite{DBLP:conf/icml/NickelTK11} is the first bilinear model.
Each relation is represented by an \textit{n}-by-\textit{n} matrix and the score of triple $(h,r,t)$ is calculated by a bilinear map that corresponds to the matrix of the relation $r$ and whose arguments are $\boldsymbol{h}$ and $\boldsymbol{t}$.
Hence, RESCAL is also the most generalized bilinear model.

Extensions of RESCAL have been proposed by restricting bilinear functions.
DistMult \cite{DBLP:journals/corr/YangYHGD14a} restricts the matrices representing relations to diagonal matrices.
DistMult makes the model easy to train and eliminates the redundancy.
However, it also has the problem that the scores of $(h,r,t)$ and $(t,r,h)$ are the same.
To solve this problem, ComplEx \cite{DBLP:conf/icml/TrouillonWRGB16} uses complex numbers instead of real numbers and takes the conjugate of the embedding of the tail entity before calculating the bilinear map. The score of the triple is the real part of the output of the bilinear map.

Bilinear models have more redundancy than translation-based models and so easily become overfitted. 
Hence, embedding spaces are limited to low-dimensional space.
This might be a problem in a huge knowledge graph that contains large numbers of entities, because high-dimensional space is required to embed the entities so that they are adequately distinguished.

\subsection{Neural Network-based Models}
Neural network-based models have layers and an activation function like a neural network.
Neural Tensor Network (NTN) \cite{DBLP:conf/nips/SocherCMN13} has a standard linear neural network structure and a bilinear tensor structure.
This can be considered as a generalization of RESCAL.
The weight of the network is trained for each relation.
ER-MLP \cite{DBLP:conf/kdd/0001GHHLMSSZ14} is a simplified version of NTN.

Neural network-based models are the most expressive models among the three categories because they have a large number of parameters.
Hence, they can possibly capture many kinds of relations but, at the same time, they tend to overfit training data the most easily. 

\section{TransE and Its Flaw}
In this section, we explain TransE \cite{DBLP:conf/nips/BordesUGWY13} in detail and show its regularization flaw. 
In the latter part of this paper, we propose a novel model that employs a similar strategy to TransE that overcomes the flaw.

The algorithm of TransE consists of three main parts as follows:
\begin{itemize}
	\item \textit{Principle}: TransE learns embeddings so that $\boldsymbol{h}+\boldsymbol{r}=\boldsymbol{t}$ holds if $(h,r,t) \in \Delta$, where $\Delta$ denotes the set of true triples. 
	To measure how much a triple embedding follows the principle, a scoring function $f$ is used.
	Usually $L_1$ or the square of the $L_2$ norm of $\boldsymbol{h}+\boldsymbol{r}-\boldsymbol{t}$ is used as $f(h,r,t)$.
	In this case, $f(h,r,t)=0$ means $\boldsymbol{h}+\boldsymbol{r}=\boldsymbol{t}$ holds completely.
	\item \textit{Negative Sampling}: With only the principle, TransE learns the trivial solution that all entity embeddings are the same and all relation embeddings are $0$. 
	Hence, negative triples are required.
	Usually a knowledge graph contains only positive triples, so TransE makes a negative triple by changing the head or the tail entity at random for each true triple.
	This is called negative sampling.
	TransE learns embeddings so that $f(h',r,t')$ gets larger if $(h',r,t') \in \Delta '_{(h,r,t)}$, where $(h,r,t) \in \Delta$ and $\Delta '_{(h,r,t)}=\{(h',r,t)|h'\in E,\ h'\neq h\} \cup \{(h,r,t')|t'\in E, \ t'\neq t\}$.
	\item \textit{Regularization}: To not allow embeddings to diverge unlimitedly, regularization is needed.
	TransE employs normalization as regularization.
	Embeddings of entities are normalized so that their magnitude becomes 1 in each step of learning. That is, for every entity $e \in E$, $\boldsymbol{e} \in S^{n-1} \subset \mathbb{R}^n $, where $S^{n-1}$ is an \textit{n-1} dimensional sphere.
	
\end{itemize}
TransE exploits margin loss. The objective function is defined as follows:
\begin{equation}
\mathcal{L}=\sum _{(h,r,t) \in \Delta} \sum _{(h', r, t')\in \Delta '_{(h,r,t)}}[\gamma+f(h,r,t)-f(h',r,t')]_+
\end{equation}
where $[x]_+$ denotes the positive part of $x$ and $\gamma > 0$ is a margin hyperparameter. TransE is trained by using stochastic gradient descent.

\begin{figure}[tb]
	\centering
	\includegraphics[width=0.4\textwidth]{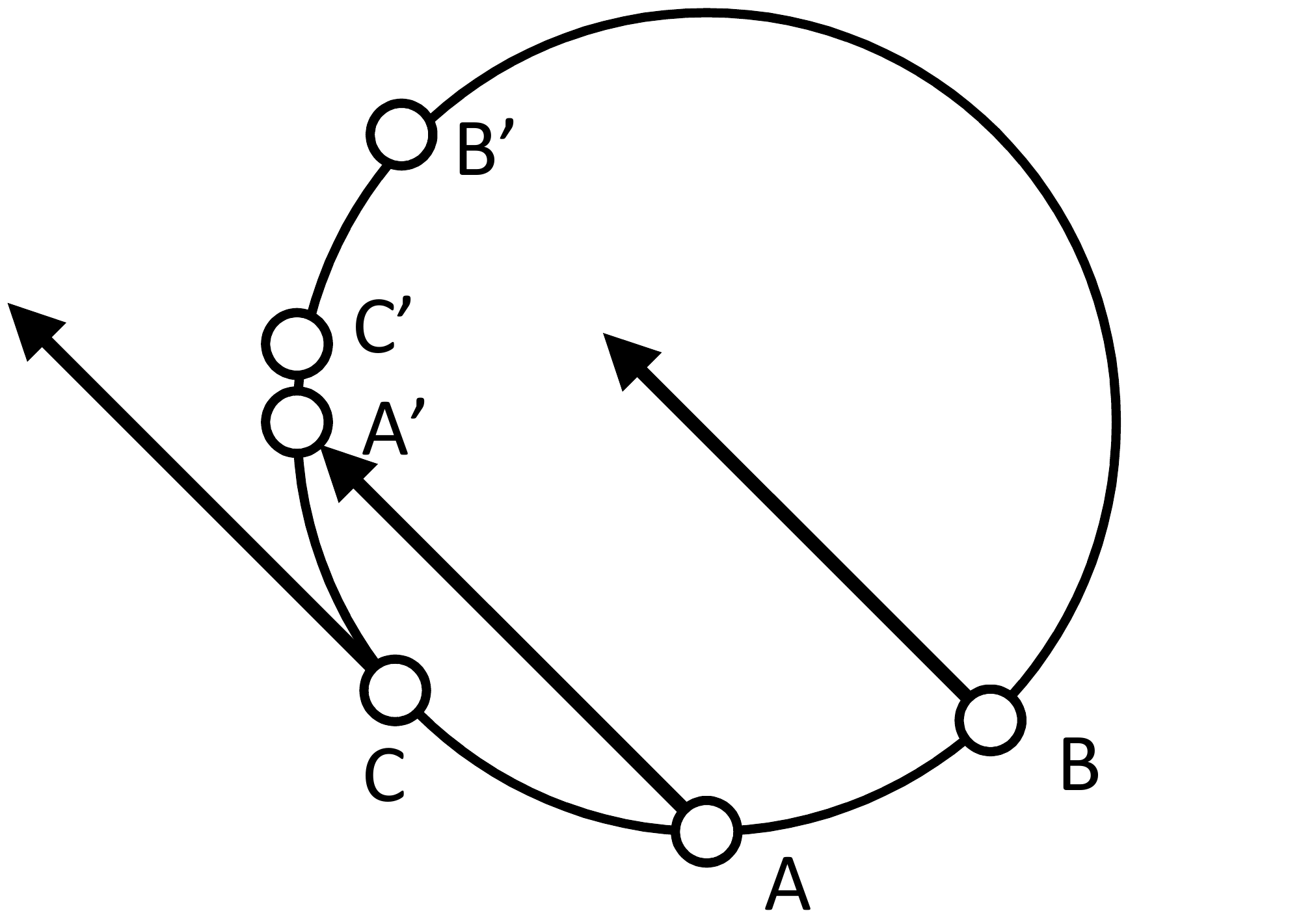}
	\caption{The image of embeddings obtained by TransE when $n$ is $2$.
		It is assumed that $(A,r,A'),(B,r,B')$ and $(C,r,C')$ hold.
	}
	\label{figure1}
\end{figure}

All three parts are necessary if entities and relations are embedded on a real vector space.
However, the principle and regularization conflict during training, because for each $e \in E$ and $r \in R$, $\boldsymbol{e}+\boldsymbol{r} \not\in S^{n-1}$ almost always holds.
Hence, the principle $\boldsymbol{h}+\boldsymbol{r}=\boldsymbol{t}$ is rarely realized in most cases, as shown in Figure \ref{figure1}.
In this figure, it is assumed that $(A,r,A'), (B,r,B')$ and $(C,r,C')$ hold.
The points represent the entity embeddings  and the arrows represent the embedding of $r$.
Embeddings of $(A,r,A')$ are obtained so that they follow the principle completely.
However, $\boldsymbol{B}+\boldsymbol{r}$ and $\boldsymbol{C}+\boldsymbol{r}$ are out of the sphere and $\boldsymbol{B}'$ and $\boldsymbol{C}'$ are regularized on it.
The regularization warps embeddings and they do not satisfy the principle. As a result, it becomes difficult to predict new triples more accurately.

\section{TorusE}
In this section, our aim is to change the embedding space to solve the regularization problem while employing the same principle used in TransE.
We first consider the required conditions for an embedding space.
Then, a Lie group is introduced as candidate embedding spaces.
After that, we propose the novel model, TorusE, which embeds entities and relations without any regularization on a torus. The torus is a compact Lie group.

\subsection{Required Conditions for Embedding Space}
To avoid the problem of regularization shown in Figure \ref{figure1}, we need to change the embedding space from $\mathbb{R}^n$, which is an open manifold, to a compact space, because  any real value continuous functions on a compact space are bounded.
It means embeddings never diverge unlimitedly because the scoring function is also bounded.
This allows us to avoid regularization and solve the conflict between the principle and the regularization during training. 
Some conditions are required for an embedding space according to the embedding strategy of TransE.
We list them as follows.
\begin{itemize}
	\item \textit{Differentiability}: The model is trained by gradient descent so that the object function is required to be differentiable. Hence, an embedding space has to be a differentiable manifold.
	\item \textit{Calculation possibility}: It is required that the principle can be defined on an embedding space.
	To do so, an embedding space has to be equipped with operations such as summation and subtraction.
	Hence, an embeddings space needs to be an Abelian group and the group operation has to be differentiable.
	\item \textit{Definability of a scoring function}: To construct an objective function for training the model, a scoring function is required to be defined on it.  
\end{itemize}
If a space fills these three conditions and is compact, we can use it as an embedding space and solve the regularization flaw of TransE.
Actually, an Abelian Lie group fills all conditions required for embedding spaces with the TransE strategy.
We explain the Lie group in the next section.
\subsection{A Lie Group}
The foundation of the theory of Lie groups was established by Sophus Lie. Lie groups, which play various roles in physics and mathematics, are defined as follows.

\begin{definition}
	A Lie group is a group that is also a finite-dimensional smooth manifold, in which the group operations of multiplication and inversion are smooth maps.
\end{definition}
A Lie group is called an Abelian Lie group when the operation of multiplication is commutative.
For an Abelian Lie group, we denote $\mu(x,y)$, $\mu(x,y^{-1})$ and $x^{-1}$ by $x+y$, $x-y$ and $-x$, respectively, where $\mu$ is the group operation.

An Abelian Lie group satisfies the \textit{Differentiability} and \textit{Calculation possibility} conditions from the definition. It is also known that distance function $d$ can be defined on any manifold.
By defining a scoring function $f(h,r,t)=d(h+r,t)$, an Abelian Lie group also satisfies the \textit{Definability}.
A real vector space as an embedding space of TransE is an example of an Abelian Lie group, because it is
a manifold and an Abelian group with ordinary vector addition as the group operation.
TransE also uses the distance function as the scoring functions derived from the norms as the vector space. However, TransE requires regularization because the real vector space is not compact.
\subsection{A Torus}
We show any Abelian Lie group can be used as an embedding space for the translation-based strategy.
We introduce a torus, which is a compact Abelian Lie group, and define distance functions on the torus.
The definition of a torus is as follows. 
\begin{definition}
	An \textit{n}-dimensional torus $T^n$ is a quotient space, $\mathbb{R}^n /\sim=\{[\boldsymbol{x}]|\boldsymbol{x}\in\mathbb{R}^n\}=\{\{\boldsymbol{y}\in\mathbb{R}^n|\boldsymbol{y}\sim \boldsymbol{x}\}|\boldsymbol{x}\in\mathbb{R}^n\}$, where $\sim$ is an equivalence relation and $\boldsymbol{y}\sim \boldsymbol{x}$ if and only if $\boldsymbol{y}-\boldsymbol{x}\in \mathbb{Z}^n$.
\end{definition}
Through the natural projection $\pi : \mathbb{R}^n \to T^n,\boldsymbol{x} \mapsto [\boldsymbol{x}]$, the topology and the differential structure of a torus is derived from the vector space.
Note that $g: T^n \to S^n \subset \mathbb{C}^n, [\boldsymbol{x}] \mapsto exp(2\pi i\boldsymbol{x})$ is a diffeomorphism and $T^n$ is diffeomorphic to $\underbrace{S^1 \times S^1 \times \cdots \times S^1}_{n}$. The group operation $\mu$ is also derived from the original vector space: $\mu ([\boldsymbol{x}],[\boldsymbol{y}])=[\boldsymbol{x}]+[\boldsymbol{y}]\triangleq [\boldsymbol{x}+\boldsymbol{y}]$. A torus is a compact Abelian Lie group with these structures and group operation.
We define distance functions in three ways: 
\begin{itemize}
	\item $d_{L_1}$: A distance function $d_{L_1}$ on $T^n$ is derived from the $L_1$ norm of the original vector space by defining $d_{L_1}([\boldsymbol{x}],[\boldsymbol{y}])=\min_{(\boldsymbol{x}',\boldsymbol{y}')\in [\boldsymbol{x}]\times [\boldsymbol{y}]} ||\boldsymbol{x}'-\boldsymbol{y}'||_1$.
\item $d_{L_2}$: A distance function $d_{L_2}$ on $T^n$ is derived from the $L_2$ norm of the original vector space by defining $d_{L_2}([\boldsymbol{x}],[\boldsymbol{y}])=\min_{(\boldsymbol{x}',\boldsymbol{y}')\in [\boldsymbol{x}]\times [\boldsymbol{y}]} ||\boldsymbol{x}'-\boldsymbol{y}'||_2$.
\item $d_{eL_2}$: $T^n$ can be embedded on $\mathbb{C}^n$ by $g$.
A distance function $d_{eL_2}$ on $T^n$ is derived from the $L_2$ norm of the $\mathbb{C}^n$ by defining $d_{eL_2}([\textbf{x}],[\textbf{y}])=||g([\textbf{x}])-g([\textbf{y}])||_2$.
\end{itemize}
These distance functions are used to define scoring functions for our model shown in the following section.
%These scoring functions are normalized so that their maximam value is $n$.
%These scoring functions and their derivatives when $n=1$ are illustrated in fig 2.
%$f_{L_1}$, $f_{L_2}$ and $f_{eL_2}$ look similar, however their derivatives are surprisingly defferent.
%$f'_{L_1}$ is constant, $f'_{L_2}$ has a vanishing point where $x=0$ and $f'_{eL_2}$ has two vanishing points where $x=0.5$, the farest point from $0$ additionaly.
%These affect obtained embeddings through gradient descent learning.
\subsection{TorusE}
\begin{figure}[tb]
	\begin{tabular}{cc}
		\begin{minipage}{0.45\columnwidth}
			\includegraphics[width=3.8cm]{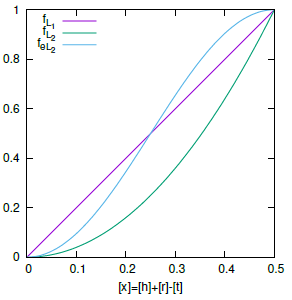}
		\end{minipage}
		&
		
		\begin{minipage}{0.45\columnwidth}
			\includegraphics[width=3.8cm]{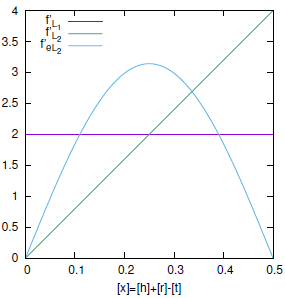}
		\end{minipage}\\
	\end{tabular}
	\caption{The graphs of scoring functions and their derivatives for TorusE when $n=1$.
		$f_{L_1}'$, $f_{L_2}'$ and $f_{eL_2}'$ are derivatives of the scoring functions. 
	}
	\label{figure2}
\end{figure}

\begin{table*}[t]
	\caption{Scoring functions for triple $(h,r,t)$, parameters and complexity of related work.}
	\label{scoring functions}
	\begin{center}
		\scalebox{1}{
			\begin{tabular}{|c|c|c|c|c|} \hline
				Model&Scoring Function&Parameters&$\mathcal{O}_{time}$&$\mathcal{O}_{space}$\\\hline\hline
				TransE&	$||\boldsymbol{h}+\boldsymbol{r}-\boldsymbol{t}||_i$&$\boldsymbol{h},\boldsymbol{r},\boldsymbol{t}\in \mathbb{R}^n$&$\mathcal{O}(n)$&	$\mathcal{O}(n)$\\\hline
				TransH& $||(\boldsymbol{h}-\boldsymbol{w}_r^\mathrm{T} \boldsymbol{h} \boldsymbol{w}_r)+\boldsymbol{r}-(\boldsymbol{t}-\boldsymbol{w}_r^\mathrm{T} \boldsymbol{t} \boldsymbol{w}_r)||_i$	&$\boldsymbol{h},\boldsymbol{r},\boldsymbol{t},\boldsymbol{w}_r\in \mathbb{R}^n
				$&$\mathcal{O}(n)$&	$\mathcal{O}(n)$\\\hline
				TransR&	$||\boldsymbol{W}_r\boldsymbol{h}+\boldsymbol{r}-\boldsymbol{W}_r\boldsymbol{t}||_i$&$\boldsymbol{h},\boldsymbol{t}\in \mathbb{R}^n, \boldsymbol{r}\in \mathbb{R}^k, \boldsymbol{W}_r \in \mathbb{R}^{k \times n}$&$\mathcal{O}(kn)$&	$\mathcal{O}(kn)$\\\hline				
				RESCAL&	$\boldsymbol{h}^\mathrm{T}\boldsymbol{W}_r\boldsymbol{t}$&$\boldsymbol{h},\boldsymbol{t}\in \mathbb{R}^n, \boldsymbol{W}_r \in \mathbb{R}^{n \times n}$&$\mathcal{O}(n^2)$&	$\mathcal{O}(n^2)$\\\hline
				DistMult&	$\boldsymbol{h}^\mathrm{T}diag(\boldsymbol{r})\boldsymbol{t}$&$\boldsymbol{h},\boldsymbol{t},\boldsymbol{r}\in \mathbb{R}^n$&$\mathcal{O}(n)$&	$\mathcal{O}(n)$\\\hline
				ComplEx&	$Re(\boldsymbol{h}^\mathrm{T}diag(\boldsymbol{r})\overline{\boldsymbol{t}})$&$\boldsymbol{h},\boldsymbol{t},\boldsymbol{r}\in \mathbb{C}^n$&$\mathcal{O}(n)$&	$\mathcal{O}(n)$\\\hline
				NTN&	$\boldsymbol{u}_r^\mathrm{T}tanh(\boldsymbol{h}^\mathrm{T}\boldsymbol{W}_r^{[1:k]}\boldsymbol{t}+\boldsymbol{V}_{r,h}\boldsymbol{h}+\boldsymbol{V}_{r,t}\boldsymbol{t}+\boldsymbol{b}_r)$&
				\shortstack{$\boldsymbol{h},\boldsymbol{t}\in \mathbb{R}^n, \boldsymbol{u}_r,\boldsymbol{b}_r \in \mathbb{R}^k, $\\$\boldsymbol{W}_r^{[1:k]}\in \mathbb{R}^{k\times n\times n}, \boldsymbol{V}_{r,h},\boldsymbol{V}_{r,t}\in \mathbb{R}^{k \times n}$}&
				$\mathcal{O}(kn^2)$&	$\mathcal{O}(kn^2)$\\\hline
				TorusE&	$min_{(\boldsymbol{x},\boldsymbol{y})\in ([\boldsymbol{h}]+[\boldsymbol{r}])\times [\boldsymbol{t}] }||\boldsymbol{x}-\boldsymbol{y}||_i$&$[\boldsymbol{h}],[\boldsymbol{r}],[\boldsymbol{t}]\in T^n$&$\mathcal{O}(n)$&	$\mathcal{O}(n)$\\\hline
				
			\end{tabular}
		}
	\end{center}
\end{table*}
TransE assumes embeddings of entities and relations on $\mathbb{R}^n$.
If $(h,r,t)$ holds for TransE, embeddings should follow the principle $\boldsymbol{h}+\boldsymbol{r}=\boldsymbol{t}$; otherwise, $\boldsymbol{h}+\boldsymbol{r}$ should be far away from $\boldsymbol{t}$.
Our proposed method, TorusE, follows the principle also, but the embedding space is changed from a vector space to a torus.
To explain the strategy, we define scoring functions in three ways that exploit the distance functions described in the previous section:
\begin{itemize}
	\item $f_{L_1}$: We define a scoring function $f_{L_1}(h,r,t)$ as $2d_{L_1}([\boldsymbol{h}]+[\boldsymbol{r}],[\boldsymbol{t}])$.
	\item $f_{L_2}$: We define a scoring function $f_{L_2}(h,r,t)$ as $4d_{L_2}^2([\boldsymbol{h}]+[\boldsymbol{r}],[\boldsymbol{t}])$.
	\item $f_{eL_2}$: We define a scoring function $f_{eL_2}(h,r,t)$ as $d_{eL_2}^2([\boldsymbol{h}]+[\boldsymbol{r}],[\boldsymbol{t}])/4$.
\end{itemize}
These scoring functions are normalized so that their maximum values are $n$.
These scoring functions and their derivatives when $n=1$ are illustrated in Figure \ref{figure2}.
\begin{comment}
\begin{figure*}[tb]
	\begin{tabular}{cc}
		\begin{minipage}{1\columnwidth}
			\includegraphics[width=5cm]{scorings.PNG}
		\end{minipage}
		&
		
		\begin{minipage}{1\columnwidth}
			\includegraphics[width=5cm]{derivatives.PNG}
		\end{minipage}\\
	\end{tabular}
	\caption{The graphs of scoring functions and their derivatives for TorusE when $n=1$.
		$f_{L_1}'$, $f_{L_2}'$ and $f_{eL_2}'$ are derivatives of the scoring functions. 
	}
	\label{figure2}
\end{figure*}
\end{comment}
$f_{L_1}$, $f_{L_2}$ and $f_{eL_2}$ look similar; however their derivatives are surprisingly different.
$f'_{L_1}$ is constant, $f'_{L_2}$ has a vanishing point at $x=0$, and $f'_{eL_2}$ has two vanishing points at $x=0$ and $x=0.5$. 
These affect the obtained embeddings through gradient descent learning.

\begin{figure}[tb]
	
	\begin{center}
		\includegraphics[width=7cm]{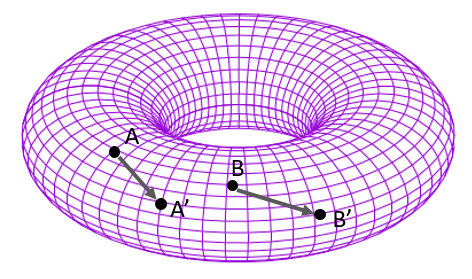}
	\end{center}

	\caption{The image of embeddings on 2-dimensional torus obtained by TorusE.
		Embeddings of the triples $(A,r,A')$ and $(B,r,B')$ are illustrated. 
		Note that $[\boldsymbol{A}']-[\boldsymbol{A}]$ and $[\boldsymbol{B}']-[\boldsymbol{B}]$ are similar on the torus.
	}
	\label{imageoftoruse}
\end{figure}

For TorusE, each entity $e \in E$ and each relation $r \in E$ are represented by $[\boldsymbol{e}]\in T^n$ and $[\boldsymbol{r}]\in T^n$, respectively.
Then, the principle is rewritten as follows:
\begin{equation}
[\boldsymbol{h}]+[\boldsymbol{r}]=[\boldsymbol{t}]
\end{equation}
and embeddings are obtained by minimizing the following objective function:
\begin{equation}
\mathcal{L}=\sum _{(h,r,t) \in \Delta} \sum _{(h', r, t')\in \Delta '_{(h,r,t)}}[\gamma+f_d(h,r,t)-f_d(h',r,t')]_+
\end{equation}
where $[x]_+$ denotes the positive part of $x$, $\gamma > 0$ is a margin hyperparameter and $f_d \in \{f_{L_1},f_{L_2},f_{eL_2}\}$.
TorusE does not require any regularization and calculation time for regularization, so it is expected to be more scalable than TransE.
The image of embeddings obtained by TorusE are shown in Figure \ref{imageoftoruse}.

The scoring functions and the complexity of related models are listed in Table \ref{scoring functions}.
Although ComplEx is a bilinear model and TorusE is a translation-based model, they have strong similarity.
By mapping $[\boldsymbol{h}], [\boldsymbol{r}]$ and $[\boldsymbol{t}]$ on $\mathbb{C}^n$ by $g$ and identifying $g([\boldsymbol{r}])$ as a corresponding diagonal matrix, $-2f_{eL_2}(h,r,t)+1=f_{ComplEx}(h,r,t)$ holds.
Bilinear models are trained to maximize the scores of triples while translation-based models are trained to minimize them.
Hence, TorusE with $f_{eL_2}$ can be considered as a more restricted and less redundant version of ComplEx on $T^n \subset \mathbb{C}^n$. 

Note that some extensions of TransE, such as TransG and pTransE, can be applied directly to TorusE by changing the embedding space from a real vector space to a torus. 
\subsubsection{Calculation Technique of a Torus}
Each embedding is represented by a point on a torus $[\boldsymbol{x}]$.
Note that $\boldsymbol{x}$ itself is an \textit{n}-dimensional vector and we use it to represent a point of the torus on a computer.
By taking a fractional part of a vector, an embedding becomes one to one with a point of the torus and we can calculate the scoring functions.
For example, we show the calculation procedure of $d_{L_1}$. 
Let $\pi _{frac} : \mathbb{R} \to [0,1)$ be the function taking a fractional part.
Then, the distance is calculated as follows:
\begin{multline*}
d_{L_1}([\boldsymbol{x}],[\boldsymbol{y}])=\\
\sum _{i=1}^n min(|\pi _{frac}(x_i)-\pi _{frac}(y_i)|,1-|\pi _{frac}(x_i)-\pi _{frac}(y_i)|)
\end{multline*}
For example, let $\boldsymbol{x}$ and $\boldsymbol{y}$ be $3.01$ and $0.99\in \mathbb{R}^1$.
Then $|\pi _{frac}(x_1)-\pi _{frac}(y_1)|=0.98$ and $1-|\pi _{frac}(x_i)-\pi _{frac}(y_i)|=0.02$ hold.
Hence we obtain $d_{L_1}([\boldsymbol{x}],[\boldsymbol{y}])=0.02$
Other distance functions are calculated in a similar way.
\section{Experiments}
We evaluated TorusE from two perspectives: one is its scalability and the other is the accuracies of the link prediction tasks.
\subsection{Datasets}
The experiments are conducted on two benchmark datasets: WN18 and FB15K \cite{DBLP:conf/nips/BordesUGWY13}.
These datasets are respectively extracted from real knowledge graphs: WordNet \cite{Miller:1995:WLD:219717.219748} and Freebase \cite{Bollacker:2008:FCC:1376616.1376746}. 
Many researchers use these datasets to evaluate models for knowledge graph completion.
The details of the datasets are shown in Table 2.
 \begin{table}[h]
 	\caption{Statistics of the datasets.}
 	\label{Table 2}
 	\begin{center}
 		\scalebox{1}{
 			\begin{tabular}{c|ccccc}
 				Dataset&\# Ent&\# Rel&\# Train &\# Valid&\# Test\\\hline 
 				WN18&40,943&18&141,442&5,000&5,000\\
 				FB15K&14,951&1,345&483,142&50,000&59,071\\

 			\end{tabular}
 		}
 	\end{center}
 \end{table}
\subsection{Experimental Setup}
\begin{figure*}[tb]
	
	\begin{tabular}{cc}
		\begin{minipage}{0.5\hsize}
			\begin{center}
				\includegraphics[width=8cm]{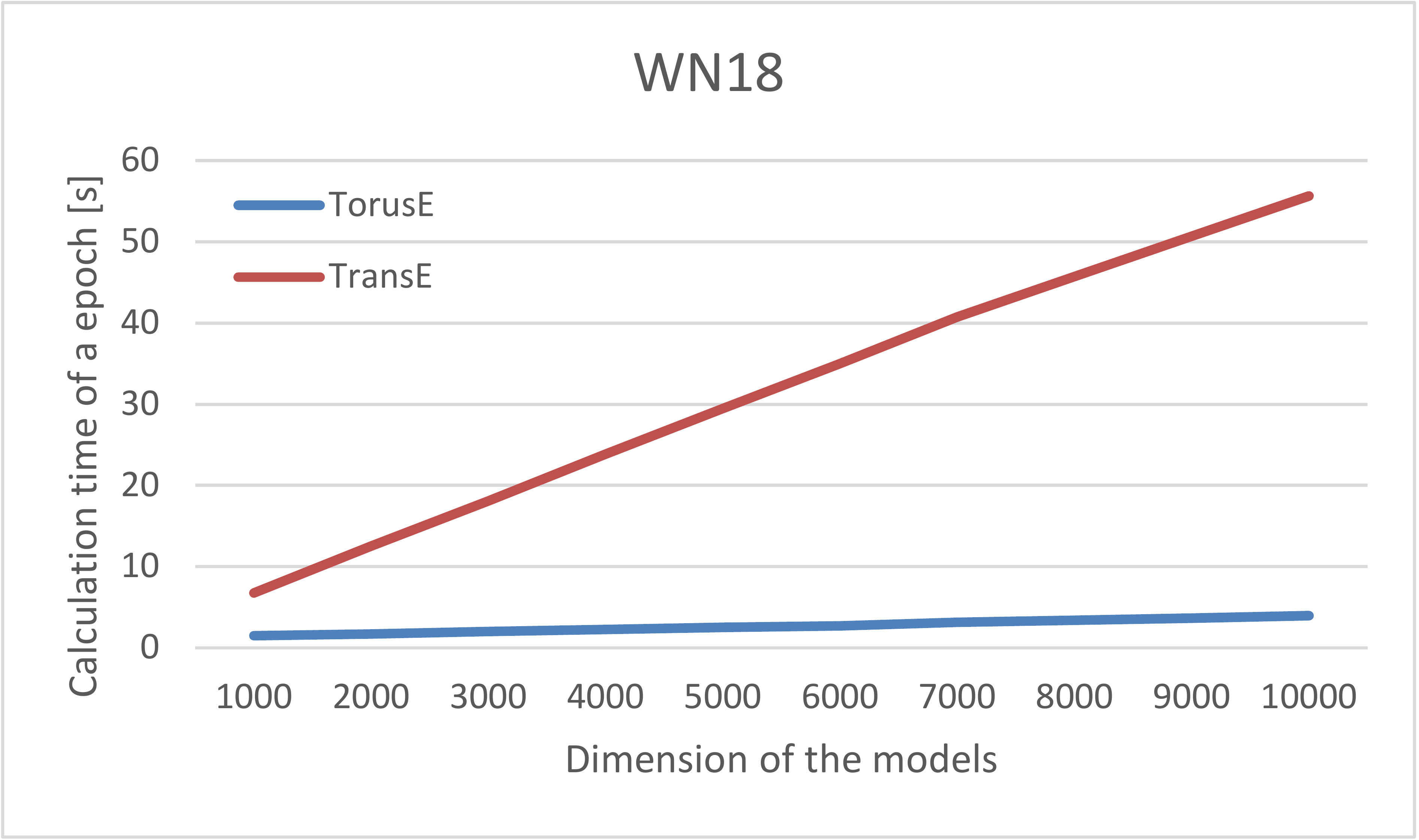}
			\end{center}
		\end{minipage}
		&
		
		\begin{minipage}{0.5\hsize}
			\includegraphics[width=8cm]{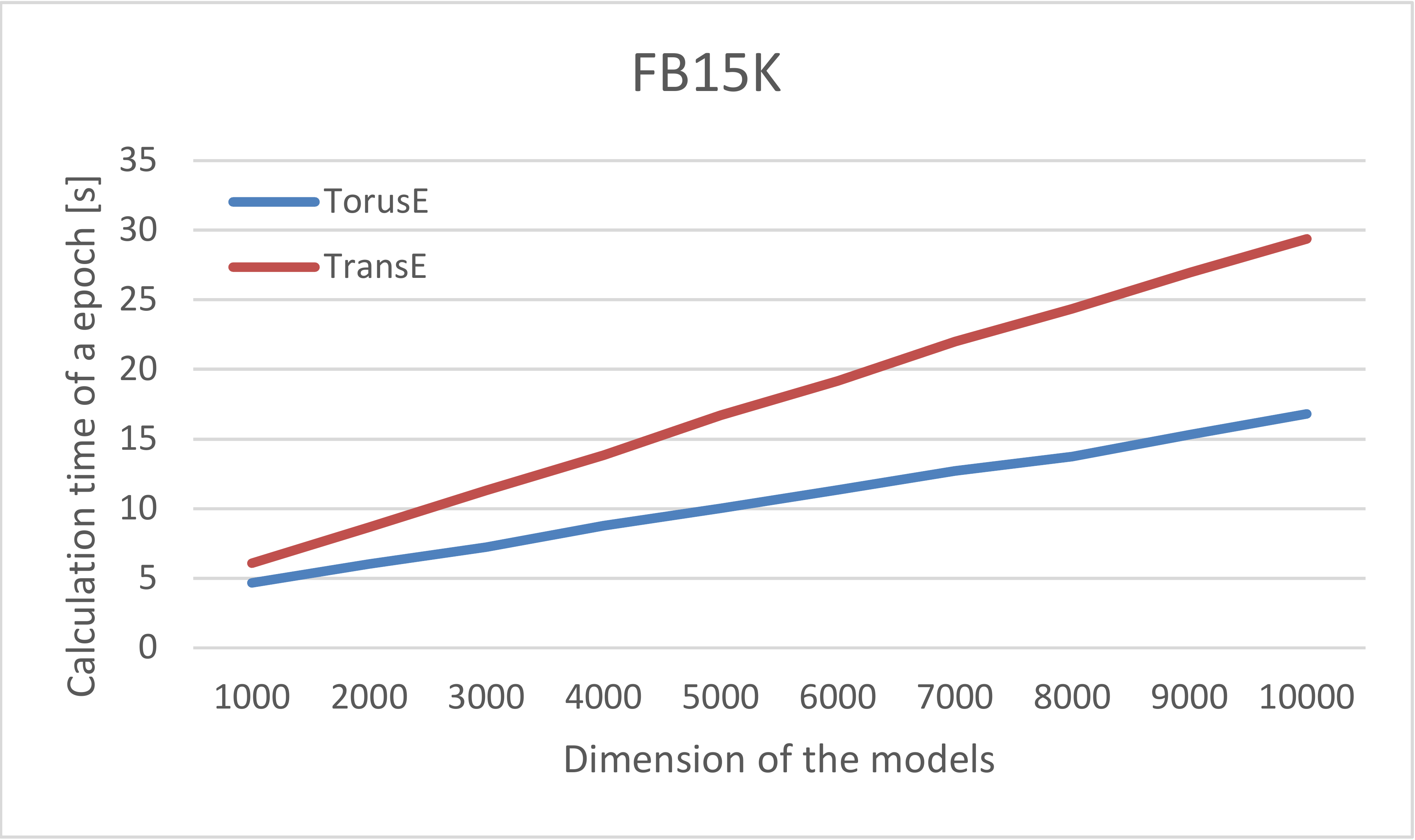}
		\end{minipage}\\
	\end{tabular}
	\caption{Calculation time of TorusE and TransE on WN18 and FB15K
	}
	\label{figure3}
\end{figure*}
\subsubsection{Evaluation Protocol}
To evaluate the scalability of TorusE, we measured the time it took to train TorusE for one epoch by changing the dimensions of the model.

We also conduct the link prediction task in the same way reported in the paper of TransE\cite{DBLP:conf/nips/BordesUGWY13}. 
For each test triple, the head (or tail) is replaced by each entity.
Then the score of each corrupted triple is calculated by the models, and the rankings of entities are obtained according to the scores. We refer to these as ``raw" rankings.
However, these rankings can be flawed when the relation and the tail have many correct entities.
In this case, the entity of the test triple might be ranked lower unfairly by other correct entities above it.
To avoid such situations as much as possible, Bordes et al. employ another ranking method, referred to as ``filtered" ranking.
A filtered ranking is obtained by eliminating entities whose corresponding triple (except the target test triple) is contained in the training, validation or test datasets.

Models are evaluated by the Mean Reciprocal Rank (MRR) and HITS@\textit{n} of these rankings.
HITS@\textit{n} is the proportion of test triples whose entity is ranked in the top \textit{n} in corresponding rankings.

%We chose TransE, TransR, RESCAL, DistMult and ComplEx as baselines.
\subsubsection{Optimization and Implementation Details}
In our implementation, TorusE was optimized by stochastic gradient descent, as for TransE.
For each epoch, we randomly separated training triples into one-hundred groups,  and embedding parameters were updated for each group.
Because the datasets contained only positive triples, we employed the "Bern"  method \cite{DBLP:conf/aaai/WangZFC14} for negative sampling.
Regularization is not required, in contrast with the other embedding methods.

We conducted a grid search to find suitable hyperparameters for each dataset.
The dimension was fixed to 10000, because a model with a higher dimension yields a better result in practice. 
We selected the margin $\gamma$ from $\{2000,1000,500,200,100\}$ and the learning rate $\alpha$ from $\{0.002,0.001,0.0005,0.0002,0.0001\}$.
Scoring functions $f_d$ were selected from $\{f_{L_1}, f_{L_2},f_{eL_2}\}$.
The best models were selected by the MRR with ``filtered" rankings on the validation set.

The optimal configurations were as follows: $\gamma = 2000, \alpha=0.0005$ and $f_d = f_{L_1}$ for WN18;
$\gamma = 500, \alpha=0.001$ and $f_d = f_{eL_2}$ for FB15K.
The results in the following section are from the models with these configurations.
\subsection{Results}
\subsubsection{Scalability of TorusE}
The calculation times of TorusE and TransE are shown in Figure \ref{figure3}.
They are measured by using a single GPU (NVIDIA Titan X).
The scoring functions of TorusE were $f_{L_1}$ for WN18 and $f_{eL_2}$ for FB15K, and the scoring functions of TransE were $L_2$ norm  for both datasets in this experiment.
The complexities of TorusE and TransE are theoretically the same and the lowest among all models at $\mathcal{O}(n)$.
For both models, the calculation time is considered a first-order equation of the dimension.
However, a large gap exists between the empirical calculation times of these models.

For the WN18 dataset, TransE takes 55.6 seconds to complete one epoch when the dimension is 10,000.
On the other hand, TorusE takes 4.0 seconds when the dimension is 10,000, and so TorusE is eleven times faster than TransE.
This is mainly due to the regularization of TransE,
because the normalizing calculations of all entity embeddings are time-consuming.

For FB15K, TransE takes 29.4 seconds to complete one epoch and TorusE takes 16.8 seconds when the dimension is 10,000, and so TorusE is faster than TransE.
FB15K contains more triples than WN18 does.
Hence, TorusE takes more time for FB15K than WN18.
However, TransE takes less time.
This is because of the number of entities contained in the datasets. The number of entities of WN18 is much more than the number of entities of FB15K.

These models were trained for 500 epochs in an experiment. So, the total time was about 30 minutes and 2 hours 30 minutes for TorusE to finish training on WN18 and FB15K respectively.
We also measured the calculation times of ComplEx with the implementation by Trouilon et al\shortcite{DBLP:journals/corr/TrouillonDWRGB17}.
The calculation times of ComplEx were 1 hour 15 minutes and 3 hours 50 minutes on each dataset with 150 and 200 dimensions on the same GPU, and TorusE was faster than it.
\subsubsection{Accuracies of the Link Prediction Tasks}
The results of the link prediction tasks are shown in Table \ref{result1}.
\begin{table*}[tb]
	\caption{Results of the link prediction tasks by Mean Reciprocal Rank (MRR) and HITS@\textit{n} on WN18 and FB15K datasets. MRRs are calculated by using the raw and filtered ranking and HITS@\textit{n} are calculated by using the filtered ranking.
		The dimension of TransE was set to 10,000, and the best hyperparameters were chosen by using the validation set.
		The results of TransR and RESCAL were reported by Nickel et al. \shortcite{DBLP:conf/aaai/NickelRP16}, the results of DistMult and ComplEx were reported by Trouillon et al. \shortcite{DBLP:conf/icml/TrouillonWRGB16}}
	\label{result1} 
	\begin{center}
		\scalebox{1}{
			\begin{tabular}{ccccccccccc} \hline
				&\multicolumn{5}{c}{WN18}&\multicolumn{5}{c}{FB15K}\\
				\cmidrule(rl){2-6}\cmidrule(l){7-11}
				&\multicolumn{2}{c}{MRR}&\multicolumn{3}{c}{HITS@}&\multicolumn{2}{c}{MRR}&\multicolumn{3}{c}{HITS@}\\
				\cmidrule(rl){2-3}\cmidrule(rl){4-6}\cmidrule(rl){7-8}\cmidrule(l){9-11}
				Model&Filtered&Raw&\textit{1}&\textit{3}&\textit{10}&Filtered&Raw&\textit{1}&\textit{3}&\textit{10}\\\hline
				%TransE&0.947&
				TransE&0.397&0.306&0.040&0.745&0.923&0.414&0.235&0.247&0.534&0.688\\
				TransR&0.605&0.427&0.335&0.876&0.940&0.346&0.198&0.218&0.404&0.582\\
				RESCAL&0.890&0.603&0.842&0.904&0.928&0.354&0.189&0.235&0.409&0.587\\
				DistMult&0.822&0.532&0.728&0.914&0.936&0.654&0.242&0.546&0.733&0.824\\
				ComplEx&0.941&0.587&0.936&0.945&0.947&0.692&0.242&0.599&0.759&\textbf{0.840}\\\hline
				TorusE&\textbf{0.947}&\textbf{0.619}&\textbf{0.943}&\textbf{0.950}&\textbf{0.954}&\textbf{0.733}&\textbf{0.256}&\textbf{0.674}&\textbf{0.771}&0.832\\\hline

			\end{tabular}
		}
	\end{center}
\end{table*}
Our method, TorusE, outperforms all other models on all metrics except HITS@\textit{10} on FB15K.
TorusE is second even on FB15K and the difference between TorusE and the best model, ComplEx, is only 0.8\%.

As is shown, TransE and extended models of TransE do not yield good results on HITS@\textit{1}, although they perform well on HITS@\textit{10}.
We believe that this phenomenon is caused by regularization of the models, even though the principle of TransE has the potential to represent real knowledge and to achieve knowledge graph completion.
We took this approach to change the embedding space in order to avoid regularization.
Therefore, TorusE can perform well on HITS@\textit{1}. 
The differences between TorusE and TransE for HITS@\textit{1} are 90.3\% on WN18 and 42.7\% on FB15K. 

Recently, bilinear models such as DistMult and ComplEx have performed far better on HITS@\textit{1}.
TorusE outperforms them also.
The accuracy of ComplEx on WN18 is already very high at 93.6\%, but the accuracy of TorusE is higher at 94.3\%.
The difference is more noticeable on FB15K.
TorusE largely outperforms on HITS@\textit{1} and yields the score of 67.4\%.
As mentioned in the section of related work, TorusE can be viewed as the restricted version of ComplEx.
Hence, TorusE has less redundancy than ComplEx.
We think this lesser redundancy accounts for the difference of accuracy.

The details of the MRR for each relation on WN18 are shown in Table \ref{details}.
\begin{table}[tb]
	\caption{Details of ``filtered" MRR on WN18. The results are listed separately for each relation contained in the dataset.}
	\label{details}
	\begin{center}
		\scalebox{1}{
			\begin{tabular}{c|ccc}
				Relation name&TorusE&ComplEx&TransE\\\hline
				hypernym&\textbf{0.957}&0.953&0.376\\
				hyponym&\textbf{0.956}&0.946&0.379\\
				member\_meronym&\textbf{0.931}&0.921&0.433\\
				member\_holonym&0.942&\textbf{0.946}&0.438\\
				instance\_hypernym&0.961&\textbf{0.965}&0.680\\
				instance\_hyponym&\textbf{0.961}&0.945&0.626\\
				has\_part&\textbf{0.944}&0.933&0.417\\
				part\_of&\textbf{0.947}&0.940&0.415\\
				member\_of\_domain\_topic&\textbf{0.944}&0.924&0.502\\
				synset\_domain\_topic\_of&0.921&\textbf{0.930}&0.536\\
				member\_of\_domain\_usage&\textbf{0.917}&\textbf{0.917}&0.270\\
				synset\_domain\_usage\_of&0.940&\textbf{1.000}&0.182\\
				member\_of\_domain\_region&\textbf{0.885}&0.865&0.358\\
				synset\_domain\_region\_of&\textbf{0.919}&\textbf{0.919}&0.197\\
				derivationally\_related\_form&\textbf{0.951}&0.946&0.362\\
				similar\_to&\textbf{1.000}&\textbf{1.000}&0.242\\
				verb\_group&\textbf{0.974}&0.936&0.283\\
				also\_see&\textbf{0.626}&0.603&0.257\\			
				
			\end{tabular}
		}
	\end{center}
\end{table}
For many relations, TorusE performs equal to or better than ComplEx.
As noted, the problem of the principle of TransE is that it cannot deal with 1-N, N-1 or N-N relations.
However, it seems TorusE can predict entities correctly for such relations, even though it employs the same principle as TransE.
We think this is because the principle itself is not actually problematic.
It is definitely impossible to follow the principle completely on such relations, but to follow the principle completely is not necessary to deal with the link prediction task.
Because the task employs rankings of entities, a model for the task is adequate when the correct entities are located in higher ranks than the incorrect entities are, even if the correct entities are not at the top ranks.

We did not conduct a grid search with changing dimensions of TorusE.
Because a preparatory experiment showed that a higher dimension seemed to yield a better result for the TorusE model. 
We think these results occur because the principle is enough restricted and the model is hard to overfit to a dataset, and the high-dimensional embedding space allows the model to represent embeddings more richly. 
This gives us a good guideline to set the hyperparameters. We set the dimension as high as possible and then find the optimal margin and the optimal learning rate.

\section{Conclusions and Future Work}
Our contributions in this paper are as follows.
\begin{itemize}
	\item We pointed out the problem of TransE: regularization.
	Regularization conflicts with the principle and makes the accuracy of the link prediction task lower.
	\item To solve this problem, we aimed to change the embedding space by using the same principle as TransE.
	By embedding on a compact space, regularization is no longer required.
	The required condition for an embedding space was clarified by finding a suitable space.
	\item We showed that a Lie group fills all conditions required.
	Then, we introduced a torus, which is a compact Lie group that can be easily realized.
	\item We proposed the novel model, TorusE, which is a model that embeds entities and relations on a torus.
	Unlike other models, it does not employ any regularization for embeddings.
	TorusE outperformed state-of-the-art models for link prediction tasks on the WN18 and FB15K datasets and it was experimentally shown to be faster than TransE.
\end{itemize}	
	
	In future work, we will consider other embedding spaces, because we only employed a torus, even though we showed all Lie groups can be used as an embedding space.
	As another approach, we will try to combine TorusE with other extended models of TransE.  
	Some of these models can be directly applied to TorusE by changing an embedding space from a vector space to a torus.
	
	Moreover, we have to consider more general models to complete a knowledge graph which can retrieve information from other materials than triples, because sometimes information is not included training triples to predict a required triple.
	There are models extracting triples from text such as OpenIE models \cite{Fader:2011:IRO:2145432.2145596,Mausam:2012:OLL:2390948.2391009,conf/acl/AngeliPM15} and word embedding-based model \cite{EBISU}.
	We think we can develop a more general model combining with these methods.
	
\section*{Acknowledgements}
This work was partially supported by the New Energy and Industrial Technology Development Organization (NEDO).
\bibliographystyle{aaai}
\bibliography{bib}

\begin{thebibliography}{}

\bibitem[\protect\citeauthoryear{Angeli, Premkumar, and
  Manning}{2015}]{conf/acl/AngeliPM15}
Angeli, G.; Premkumar, M. J.~J.; and Manning, C.~D.
\newblock 2015.
\newblock Leveraging linguistic structure for open domain information
  extraction.
\newblock In {\em Proceedings of the 53rd Annual Meeting of the Association for
  Computational Linguistics},  344--354.

\bibitem[\protect\citeauthoryear{Auer \bgroup et al\mbox.\egroup
  }{2007}]{DBLP:conf/semweb/AuerBKLCI07}
Auer, S.; Bizer, C.; Kobilarov, G.; Lehmann, J.; Cyganiak, R.; and Ives, Z.~G.
\newblock 2007.
\newblock {DB}pedia: {A} nucleus for a web of open data.
\newblock In {\em The Semantic Web, 6th International Semantic Web Conference,
  2nd Asian Semantic Web Conference},  722--735.

\bibitem[\protect\citeauthoryear{Bollacker \bgroup et al\mbox.\egroup
  }{2008}]{Bollacker:2008:FCC:1376616.1376746}
Bollacker, K.; Evans, C.; Paritosh, P.; Sturge, T.; and Taylor, J.
\newblock 2008.
\newblock Freebase: A collaboratively created graph database for structuring
  human knowledge.
\newblock In {\em Proceedings of the 2008 ACM SIGMOD International Conference
  on Management of Data},  1247--1250.

\bibitem[\protect\citeauthoryear{Bordes \bgroup et al\mbox.\egroup
  }{2013}]{DBLP:conf/nips/BordesUGWY13}
Bordes, A.; Usunier, N.; Garc{\'{\i}}a{-}Dur{\'{a}}n, A.; Weston, J.; and
  Yakhnenko, O.
\newblock 2013.
\newblock Translating embeddings for modeling multi-relational data.
\newblock In {\em Advances in Neural Information Processing Systems},
  2787--2795.

\bibitem[\protect\citeauthoryear{Dong \bgroup et al\mbox.\egroup
  }{2014}]{DBLP:conf/kdd/0001GHHLMSSZ14}
Dong, X.; Gabrilovich, E.; Heitz, G.; Horn, W.; Lao, N.; Murphy, K.; Strohmann,
  T.; Sun, S.; and Zhang, W.
\newblock 2014.
\newblock Knowledge vault: a web-scale approach to probabilistic knowledge
  fusion.
\newblock In {\em Proceedings of The 20th {ACM} {SIGKDD} International
  Conference on Knowledge Discovery and Data Mining},  601--610.

\bibitem[\protect\citeauthoryear{Ebisu and Ichise}{2017}]{EBISU}
Ebisu, T., and Ichise, R.
\newblock 2017.
\newblock Triple prediction from texts by using distributed representations of
  words.
\newblock {\em IEICE Transactions on Information and Systems}
  Vol.E100-D(12):3001--3009.

\bibitem[\protect\citeauthoryear{Fader, Soderland, and
  Etzioni}{2011}]{Fader:2011:IRO:2145432.2145596}
Fader, A.; Soderland, S.; and Etzioni, O.
\newblock 2011.
\newblock Identifying relations for open information extraction.
\newblock In {\em Proceedings of the Conference on Empirical Methods in Natural
  Language Processing},  1535--1545.

\bibitem[\protect\citeauthoryear{Lin \bgroup et al\mbox.\egroup
  }{2015a}]{DBLP:conf/emnlp/LinLLSRL15}
Lin, Y.; Liu, Z.; Luan, H.; Sun, M.; Rao, S.; and Liu, S.
\newblock 2015a.
\newblock Modeling relation paths for representation learning of knowledge
  bases.
\newblock In {\em Proceedings of the 2015 Conference on Empirical Methods in
  Natural Language Processing},  705--714.

\bibitem[\protect\citeauthoryear{Lin \bgroup et al\mbox.\egroup
  }{2015b}]{DBLP:conf/aaai/LinLSLZ15}
Lin, Y.; Liu, Z.; Sun, M.; Liu, Y.; and Zhu, X.
\newblock 2015b.
\newblock Learning entity and relation embeddings for knowledge graph
  completion.
\newblock In {\em Proceedings of the Twenty-Ninth {AAAI} Conference on
  Artificial Intelligence},  2181--2187.

\bibitem[\protect\citeauthoryear{Mausam \bgroup et al\mbox.\egroup
  }{2012}]{Mausam:2012:OLL:2390948.2391009}
Mausam; Schmitz, M.; Bart, R.; Soderland, S.; and Etzioni, O.
\newblock 2012.
\newblock Open language learning for information extraction.
\newblock In {\em Proceedings of the 2012 Joint Conference on Empirical Methods
  in Natural Language Processing and Computational Natural Language Learning},
  523--534.

\bibitem[\protect\citeauthoryear{Mikolov \bgroup et al\mbox.\egroup
  }{2013a}]{DBLP:journals/corr/abs-1301-3781}
Mikolov, T.; Chen, K.; Corrado, G.; and Dean, J.
\newblock 2013a.
\newblock Efficient estimation of word representations in vector space.
\newblock {\em CoRR} abs/1301.3781.

\bibitem[\protect\citeauthoryear{Mikolov \bgroup et al\mbox.\egroup
  }{2013b}]{DBLP:conf/nips/MikolovSCCD13}
Mikolov, T.; Sutskever, I.; Chen, K.; Corrado, G.~S.; and Dean, J.
\newblock 2013b.
\newblock Distributed representations of words and phrases and their
  compositionality.
\newblock In {\em Advances in Neural Information Processing Systems},
  3111--3119.

\bibitem[\protect\citeauthoryear{Miller}{1995}]{Miller:1995:WLD:219717.219748}
Miller, G.~A.
\newblock 1995.
\newblock Wordnet: A lexical database for {E}nglish.
\newblock {\em Commun. ACM} 38(11):39--41.

\bibitem[\protect\citeauthoryear{Nickel, Rosasco, and
  Poggio}{2016}]{DBLP:conf/aaai/NickelRP16}
Nickel, M.; Rosasco, L.; and Poggio, T.~A.
\newblock 2016.
\newblock Holographic embeddings of knowledge graphs.
\newblock In {\em Proceedings of the Thirtieth {AAAI} Conference on Artificial
  Intelligence},  1955--1961.

\bibitem[\protect\citeauthoryear{Nickel, Tresp, and
  Kriegel}{2011}]{DBLP:conf/icml/NickelTK11}
Nickel, M.; Tresp, V.; and Kriegel, H.
\newblock 2011.
\newblock A three-way model for collective learning on multi-relational data.
\newblock In {\em Proceedings of the 28th International Conference on Machine
  Learning},  809--816.

\bibitem[\protect\citeauthoryear{Socher \bgroup et al\mbox.\egroup
  }{2013}]{DBLP:conf/nips/SocherCMN13}
Socher, R.; Chen, D.; Manning, C.~D.; and Ng, A.~Y.
\newblock 2013.
\newblock Reasoning with neural tensor networks for knowledge base completion.
\newblock In {\em Advances in Neural Information Processing Systems},
  926--934.

\bibitem[\protect\citeauthoryear{Suchanek, Kasneci, and
  Weikum}{2007}]{DBLP:conf/www/SuchanekKW07}
Suchanek, F.~M.; Kasneci, G.; and Weikum, G.
\newblock 2007.
\newblock Yago: a core of semantic knowledge.
\newblock In {\em Proceedings of the 16th International Conference on World
  Wide Web},  697--706.

\bibitem[\protect\citeauthoryear{Trouillon \bgroup et al\mbox.\egroup
  }{2016}]{DBLP:conf/icml/TrouillonWRGB16}
Trouillon, T.; Welbl, J.; Riedel, S.; Gaussier, {\'{E}}.; and Bouchard, G.
\newblock 2016.
\newblock Complex embeddings for simple link prediction.
\newblock In {\em Proceedings of the 33rd International Conference on Machine
  Learning},  2071--2080.

\bibitem[\protect\citeauthoryear{Trouillon \bgroup et al\mbox.\egroup
  }{2017}]{DBLP:journals/corr/TrouillonDWRGB17}
Trouillon, T.; Dance, C.~R.; Welbl, J.; Riedel, S.; Gaussier, {\'{E}}.; and
  Bouchard, G.
\newblock 2017.
\newblock Knowledge graph completion via complex tensor factorization.
\newblock {\em CoRR} abs/1702.06879.

\bibitem[\protect\citeauthoryear{Wang \bgroup et al\mbox.\egroup
  }{2014}]{DBLP:conf/aaai/WangZFC14}
Wang, Z.; Zhang, J.; Feng, J.; and Chen, Z.
\newblock 2014.
\newblock Knowledge graph embedding by translating on hyperplanes.
\newblock In {\em Proceedings of the Twenty-Eighth {AAAI} Conference on
  Artificial Intelligence},  1112--1119.

\bibitem[\protect\citeauthoryear{Xiao, Huang, and
  Zhu}{2016}]{DBLP:conf/acl/0005HZ16}
Xiao, H.; Huang, M.; and Zhu, X.
\newblock 2016.
\newblock Trans{G} : {A} generative model for knowledge graph embedding.
\newblock In {\em Proceedings of the 54th Annual Meeting of the Association for
  Computational Linguistics}.

\bibitem[\protect\citeauthoryear{Yang \bgroup et al\mbox.\egroup
  }{2014}]{DBLP:journals/corr/YangYHGD14a}
Yang, B.; Yih, W.; He, X.; Gao, J.; and Deng, L.
\newblock 2014.
\newblock Embedding entities and relations for learning and inference in
  knowledge bases.
\newblock {\em CoRR} abs/1412.6575.

\end{thebibliography}

\end{document}